\documentclass[11pt]{article}
\pdfoutput=1

\usepackage{authblk}
\usepackage[]{acl}

\usepackage{times}
\usepackage{latexsym}

\usepackage[T1]{fontenc}

\usepackage[utf8]{inputenc}

\usepackage{microtype}

\usepackage{inconsolata}

\usepackage{todonotes}
\usepackage{booktabs}
\usepackage{subcaption}
\usepackage{graphicx}
\usepackage{amssymb}
\usepackage[ruled,vlined,linesnumbered]{algorithm2e}
\usepackage{amsmath}
\usepackage{makecell}
\usepackage{multirow}
\usepackage{array}
\usepackage{comment}
\usepackage{hyperref}

\title{Improving Explicit Spatial Relationships in Text-to-Image Generation through an Automatically Derived Dataset}

\author[1]{\textbf{Ander Salaberria}}
\author[1]{\textbf{Gorka Azkune}}
\author[1]{\textbf{Oier Lopez de Lacalle}}
\author[1]{\textbf{Aitor Soroa}}
\author[1]{\\\textbf{Eneko Agirre}}
\author[2]{\textbf{Frank Keller}}

\affil[1]{HiTZ Center - Ixa, University of the Basque Country UPV/EHU}
\affil[ ]{\{ander.salaberria, gorka.azcune, oier.lopezdelacalle, a.soroa, e.agirre\}@ehu.eus}
\affil[2]{University of Edinburgh}
\affil[ ]{keller@inf.ed.ac.uk}

\begin{document}
\maketitle
\begin{abstract}

Existing work has observed that current text-to-image systems do not accurately reflect explicit spatial relations between objects such as \emph{left of} or \emph{below}. We hypothesize that this is because explicit spatial relations rarely appear in the image captions used to train these models. We propose an automatic method that, given existing images, generates synthetic captions that contain 14 explicit spatial relations. We introduce the Spatial Relation for Generation (SR4G) dataset, which contains 9.9 millions image-caption pairs for training, and more than 60 thousand captions for evaluation. In order to test generalization we also provide an \emph{unseen} split, where the set of objects in the train and test captions are disjoint. SR4G is the first dataset that can be used to spatially fine-tune text-to-image systems. We show that fine-tuning two different Stable Diffusion models (denoted as SD$_{SR4G}$) yields up to 9 points improvements in the VISOR metric. The improvement holds in the \emph{unseen} split, showing that SD$_{SR4G}$ is able to generalize to unseen objects. SD$_{SR4G}$ improves the state-of-the-art with fewer parameters, and avoids complex architectures. Our analysis shows that improvement is consistent for all relations. The dataset and the code are publicly available.\footnote{Url: \href{https://github.com/salanueva/sr4g}{https://github.com/salanueva/sr4g}}

\end{abstract}

\section{Introduction}

Text-to-image generators such as Midjourney, 
Stable Diffusion \cite{rombach2022high} and Dalle-3 \cite{betker2023improving} have recently made rapid advances and generated a lot of interest. However, those systems are still far from being perfect and show some important weaknesses. For instance, as observed by \cite{gokhale2023benchmarking} and \cite{cho2023visual} among others, current text-to-image generators do not represent well explicit spatial relations like \emph{left of} or \emph{below}, which limits their capabilities for important applications like text-based image editing \cite{kawar2023imagic}.

We hypothesize that the poor performance for explicit spatial relations is due to the lack of such relations in the datasets used to train those models. To support our hypothesis we analysed the LAION-2B dataset \cite{schuhmann2022laion}, which has been used to train the state-of-the-art open source model Stable Diffusion.  LAION-2B takes the captions from alt-text fields of images on the web. 
We automatically searched for explicit spatial relations (\emph{left}, \emph{right}, \emph{below} and so on) and found that only $0.72\%$ of cations contain the target words. 
Furthermore, $64.1\%$ of these relations are \emph{left} and \emph{right}, which cannot be captured by image generators, as random horizontal flips are applied to images during training. 

\begin{figure}
    \centering
    \includegraphics[width=\linewidth]{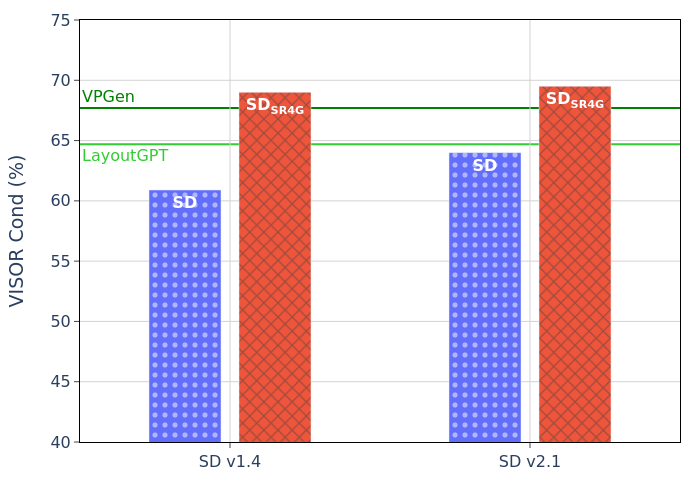}
    \caption{Fine-tuning Stable Diffusion on our SR4G dataset improves  results significantly (two versions of SD shown), surpassing the state of the art in spatial-aware systems (see Section \ref{sec:exps}).}
    \label{fig:results}
\end{figure}

Motivated by the lack of captions with spatial relations, 
we focus on the training data to improve current end-to-end diffusion models; this is complementary to proposed architectural modifications on the system itself \cite{cho2023visual, feng2023layoutgpt}. More concretely, we propose an approach to automatically generate synthetic captions which contain explicit spatial relations with paired real images. Leveraging the object annotations in COCO \cite{lin2014microsoft} and heuristic rules to infer the spatial relation between two bounding boxes, we build a dataset of real images paired with synthetic captions, called Spatial Relations for Generation (SR4G).

We use SR4G to fine-tune two Stable Diffusion models, assuming that exposure to image-caption pairs with explicit spatial relations will enhance the capabilities of the models to represent those relations. To evaluate our fine-tuned models and compare to the unmodified base models, we use the recently proposed VISOR metric \cite{gokhale2023benchmarking},
which we extend to support more spatial relations. 

The contributions of this paper are the following: (1) We release SR4G, the first benchmark that allows to fine-tune, develop and evaluate the spatial understanding capabilities of text-to-image models for 14 explicit relations; (2) Our experiments show that fine-tuning Stable Diffusion on SR4G improves the understanding of spatial relations and provides more accurate images; (3) The improvement holds even when tested on unseen objects, showing that the models are able to learn the relations, generalizing to unseen objects; (4) The results exceed the state-of-the-art in spatial understanding for image generation \cite{cho2023visual, feng2023layoutgpt} with fewer parameters and avoiding complex architectures or Large Language Models.

\section{Related Work}
Many text-to-image systems have been proposed in the last few years. In general, we can distinguish between those based on auto-regressive transformer architectures, such as the original Dall-E \cite{pmlr-v139-ramesh21a}, the multi-task system OFA \cite{wang2022ofa} or CogView2 \cite{ding2022cogview2}; and those based on diffusion models, pioneered by GLIDE \cite{nichol2022glide}, which evolved into current latent diffusion models such as Stable Diffusion \cite{rombach2022high} and Attend-and-Excite \cite{chefer2023attend}.  

Although the results of text-to-image systems keep improving, recent work has shown that their performance for explicit spatial relations is low \cite{gokhale2023benchmarking, cho2023visual}; the models struggle to correctly draw textual descriptions like \textit{a cat on top of a table}. To overcome these limitations, VPGen \cite{cho2023visual} and LayoutGPT \cite{feng2023layoutgpt} propose pipeline systems, combining Large Language Models to generate layouts from textual prompts and layout-to-image generators such as GLIGEN \cite{li2023gligen}. The difference between both systems is that VPGen fine-tunes Vicuna-13B \cite{chiang2023vicuna} to generate layouts from textual descriptions, whereas LayoutGPT relies on Llama-2-7B \cite{touvron2023llama} and in-context learning for the same purpose.\footnote{Originally they use LLMs from the OpenAI GPT family, but they have released a publicly available Llama-2 based variant of LayoutGPT, which we use in this work.}

To avoid the use of complex and large pipeline systems, \cite{yang2023reco} propose ReCo, an end-to-end system which uses layout descriptions in the input. In this paper, we also focus on end-to-end systems, but we avoid inserting layout information into the input, as this imposes a substantial burden on users compared to simple text inputs.

To evaluate the performance of text-to-image generators for explicit spatial relations, dedicated datasets have been created, since commonly used datasets like COCO \cite{lin2014microsoft}, CC12M \cite{changpinyo2021conceptual} or LAION \cite{schuhmann2022laion}, contain very few examples of explicit spatial relations. For example, \cite{gokhale2023benchmarking} propose the $SR_{2D}$ dataset, composed of synthetic captions created combining two objects in the COCO object vocabulary and four explicit spatial relations. $SR_{2D}$ only contains captions and it is thus not amenable for training. Similarly  \cite{feng2023layoutgpt} published the Numerical and Spatial Reasoning dataset (NSR-1K) which does include caption-image pairs. The spatial part contains only 1021 image-caption pairs (738 for train and 283 for test, no development) for 4 relations, insufficient for accurate evaluation and too small for training. 

Our paper proposes a new dataset with synthetic captions \textbf{and paired images} which can be used to train and evaluate spatial understanding of text-to-image generation systems, containing 14 different spatial relations and including 9.9 million image/caption pairs (Section \ref{sec:sr4g}). Finally, for evaluating the generated images, we follow \cite{gokhale2023benchmarking, feng2023layoutgpt, cho2023visual} and use an off-the-shelf object detector to extract bounding boxes and compute the spatial relation between detected objects.

\section{SR4G: A new synthetic dataset for explicit spatial relation generation}
\label{sec:sr4g}

Given the shortcomings of previous datasets, we propose to generate meaningful synthetic captions for real images, and use them to build the SR4G dataset (Spatial Relations for Generation). We increase the number of spatial relations used in previous work \cite{gokhale2023benchmarking, cho2023visual, feng2023layoutgpt} including not only projective or scale relations, but also topological ones. The full list of unambiguous spatial relations we used is as follows:

\noindent \textbf{Projective:} \emph{left of}, \emph{right of}, \emph{above} and \emph{below}.

\noindent \textbf{Topological:} \emph{overlapping}, \emph{separated}, \emph{surrounding} and \emph{inside}.

\noindent \textbf{Scale:} \emph{taller}, \emph{shorter}, \emph{wider}, \emph{narrower}, \emph{larger} and \emph{smaller}.

Our objective is to build a dataset for training, development and evaluation. For training, we need image-caption pairs, but for evaluation, captions with spatial relations are enough, since, following previous work \cite{gokhale2023benchmarking, cho2023visual}, the outputs of the image generator are not evaluated against real images. The evaluation method is described in Section \ref{sec:eval}.

\subsection{Captions for evaluation}
\label{sec:evalcaps}
We first generate a set of spatial triplets of the form $\langle$\textit{subject, relation, object}$\rangle$. We build our initial set of triplets using all pairwise combinations of the 80 objects in the vocabulary of COCO \cite{lin2014microsoft}, yielding $3,160$ object pairs, and combining each pair with all of our 14 spatial relations, resulting in 88,480 spatial triplets. 

However, some spatial triplets in the initial set are not \textit{natural}. For example, it is very difficult to find natural images for triplets like $\langle$\textit{skis, above, toothbrush}$\rangle$ or $\langle$\textit{truck, inside, cat}$\rangle$.
We want to remove those \textit{unnatural} triplets from our dataset to focus on triplets that appear in natural images. Therefore, we identify all triplets that appear at least once in the training split of the COCO dataset and use that subset to generate our evaluation captions, which consists of $68.8\%$ of the entire set of triplets (60,836 triplets).

Using hand-designed templates  to be as simple as
possible (Appendix \ref{sec:appSR4G-Templates}), we generate the final evaluation captions from the set of spatial triplets (Figure \ref{fig:qualAnalysis} shows some examples). Those captions reflect only the spatial relations between two objects, avoiding to include any other textual details.

\subsection{Image-caption pairs for training}

For training, we need captions with explicit spatial relations and real images in which those relations are depicted. We use the COCO 2017 training split to collect real images with object annotations and define a methodology to generate first spatial triplets from those images, and then textual captions derived from those triplets. 

Given an image $I$ and a list of $n$ objects $O_{I} = \{o_{1}, o_{2}, \ldots, o_{n}\}$ belonging to $I$, the goal is to generate a triplet with a valid spatial relation $r$ between two objects in $O_{I}$: $o_{s}$ and $o_{o}$, where $s, o \in \{1, \ldots, n\}$. For each object $o_{i}$, we know its respective label $l_{i}$ and bounding box (\emph{bbox}) $bb_{i} = \{x_i^0, y_i^0, x_i^1, y_i^1\}$, that is, four coordinates that define the position and size of $o_i$ in the image. 

Therefore, $t_{j} = \langle l_{s}, r, l_{o}\rangle$ is a triplet defined in SR4G that is represented in $I$. We call this set of valid triplets $T_{I} = \{t_{1}, \ldots, t_{m}\}$, where $m$ is the number of valid spatial relations in the given image $I$. This implies that each relation $r$ has to be linked to a heuristic rule $f_{r}$ where, given the \emph{bboxes} of two objects, it determines whether a given triplet is instantiated or not (see Eq. \ref{eq:heurFunc}). We follow \cite{johnson2018image} and define $f_r$ functions, which represent unambiguous spatial relations between two object bounding boxes (see Appendix \ref{sec:appSR4G-Rules}). 

\begin{equation}
    t_{j} = \langle l_{s}, r, l_{o}\rangle \in T_{I} \longleftrightarrow f_{r}(bb_{s}, bb_{o})
    \label{eq:heurFunc}
\end{equation}

We apply data augmentation strategies (random crops and horizontal flips) to the original COCO images in order to obtain an image $I$ and its object list $O_{I}$. Then, we randomly select two objects as $o_{s}$ and $o_{o}$, compute the list of valid relations using our predefined $f_{r}$ functions, and randomly select one of these relations, building the $j$-th valid relation of $I$ without computing the entire $T_{I}$ set: $t_{j} = (l_{s}, r, l_{o})$. Finally, we verbalize the obtained triplet $t_j$ using the same hand-designed templates as for the evaluation captions (Section \ref{sec:evalcaps}).

\subsection{Dataset splits}
\label{sec:splits}

\begin{table}[t]
    \centering
    \resizebox{\linewidth}{!}{
        \begin{tabular}{lccccc}
            \toprule
            \multirow{2}{*}{Splits} & \multirow{2}{*}{Images} & \multicolumn{3}{c}{Unique Captions} & \multirow{2}{*}{I/C Pairs} \\ 
            &  & Train & Val & Test &  \\ \midrule
            Main & 103.4k & 60.8k & 2.5k & 60.8k & 9.9M \\ 
            Unseen & 83.6k & 46.9k & 2.5k & 8.0k & 4.8M \\ \bottomrule
        \end{tabular}
    }
    \caption{SR4G dataset's statistics. \emph{Images} column refers to the number of images used during training, \emph{Unique triplets} column represents the amount of unique triplets, and \emph{I/C pairs} refers to the number of unique image/caption pairs that can be generated.}
    \label{tab:sr4gStats}
\end{table}

We build two different splits of SR4G, namely the \emph{main} and the \emph{unseen} splits. The \emph{main} split consists of all the spatial triplets/captions of the SR4G test set (see Section \ref{sec:evalcaps}). The training instances are generated on-the-fly without any restrictions on the triplets, which means that the same triplet can be in train, validation and test splits. For the \emph{unseen} split, we randomly divide the COCO dataset's 80 objects into training, validation and test sets of $|O_{\mathrm{train}}| = 45$, $|O_{\mathrm{val}}| = 5$ and $|O_{\mathrm{test}}| = 30$ objects, respectively. More specifically, during training we just take objects from $O_{\mathrm{train}}$ into account when randomly selecting \emph{bboxes} to dynamically build spatial captions. For validation, as there are few combinations that can be built with $O_{\mathrm{val}}$, we select triplets that contain one of these 5 objects at least once and do not contain any object that is set aside for the test split. For testing purposes we use triplets built by only using objects from $O_\mathrm{test}$. Table \ref{tab:sr4gStats} shows the relevant numbers of our splits (more details in Appendix \ref{sec:appSR4G-Splits}).

\subsection{Evaluation metrics}
\label{sec:eval}

To evaluate the performance of text-to-image systems for spatial relations, we use three evaluation metrics proposed by \cite{gokhale2023benchmarking}: 

\paragraph{Object Accuracy:} Given a generated image $I'$ and two object labels $l_{a}$ and $l_{b}$, object accuracy measures whether both objects appear in $I'$. We obtain a list of objects for $I'$, i.e., $L_{I'} = \{l_{1}, \ldots, l_{n}\}$, by using an off-the-shelf open-vocabulary object detector, OWL-ViT \cite{minderer2022simple}. This metric is useful for analyzing the object generation capabilities of an image generator, as it does not take the relation $r$ into account.

\begin{equation}
    \mathrm{OA}(I, l_{a}, l_{b}) = 
    \begin{cases}
         1  & \mathrm{if} \hspace{0.15cm} l_{a}, l_{b} \in O_{I'} \\
         0  & \mathrm{else}
   \end{cases}
   \label{eq:objAcc}
\end{equation}

\paragraph{VISOR:} Given a generated image $I'$ and a spatial triplet $t = (l_{a}, r, l_{b})$, VISOR measures whether both objects appear and if the spatial relation $r$ is valid between them. Function $f_{r}$ takes the bounding boxes of both objects ($bb_{a}$ and $bb_{b})$ and compares them to check if the triplet is valid. Bounding boxes are provided by the object detector. VISOR increases both when the model generates the requested objects and when the ratio of correctly generated relations increases, showing the ability of the model in visualising spatial triplets.
\begin{equation}
    \mathrm{VISOR}(I, t) = 
    \begin{cases}
         1  & \parbox[t]{3cm}{$\mathrm{if} \hspace{0.1cm} l_{a}, l_{b} \in L_{I'} \wedge f_{r}(bb_{a}, bb_{b})$} \\
         0  & \mathrm{else}
    \label{eq:visor}
   \end{cases}
\end{equation}
\paragraph{VISOR$_{\mathrm{Cond}}$:} This is the proportion of correctly generated spatial triplets, taking into account only images in which both objects are generated.

Given that our contribution focuses on spatial understanding, we focus on VISOR$_{\mathrm{Cond}}$, as it quantifies the ability of the model to represent spatial relations correctly without considering its object generation capability. It is the most informative measure, specially when comparing between systems which might have different object generation abilities, as it isolates the understanding of spatial relations. We thus use it as our main performance metric in the experiments, although we also report the other two metrics, while extending the number of spatial relations from~4 to~14,

\section{Experiments and Results}
\label{sec:exps}

In this section we show that end-to-end models improve their capability of depicting spatial relations when they are fine-tuned with synthetic training examples. Furthermore, we find that our fine-tuned models SD$_{SR4G}$ generalize to unseen objects during fine-tuning.

\subsection{Experimental set-up}
\label{sec:expSetUp}

\textbf{Models.} We use Stable Diffusion (SD) as the base model, as it shows the best performance on spatial relation generation among publicly available end-to-end models \cite{gokhale2023benchmarking}. 
We use two different versions of Stable Diffusion: SD v1.4 and SD v2.1, which generate images of 512x512 and 768x768 pixels, respectively. 

\textbf{Training.}  To fine-tune SD models on SR4G, we use the original loss function proposed by \cite{rombach2022high}, i.e., the mean square error over latent noise representations. 
We fine-tune SD models for 100k training steps with an effective batch-size of 64 instances, evaluating on the validation split every 5k steps. 
After training is complete, we select the checkpoint with the highest VISOR$_{\mathrm{Cond}}$ value on the validation split. Following \cite{gokhale2023benchmarking}, we generate four images per spatial relation in all of our evaluations for consistency. More details can be found in Appendixes \ref{sec:appTrain} and \ref{sec:appEval}.

\subsection{Main results} 
\label{sec:baseSplit}

\begin{table}[t]
\centering
\resizebox{\linewidth}{!}{
\begin{tabular}{lccc}
\toprule
Model & VISOR$_{\mathrm{Cond}}$ $\uparrow$ & VISOR $\uparrow$ & OA $\uparrow$ \\ \midrule
\multicolumn{4}{c}{\emph{Main split}} \\ \midrule
SD v1.4 & 60.9 & 17.6 & 29.0 \\
SD v2.1 & 64.0 & 27.4 & 42.8 \\ 
SD$_{SR4G}$ v1.4 & 69.0 & 26.8 & 38.9 \\ 
SD$_{SR4G}$ v2.1 & \textbf{69.5} & 31.7 & 45.6  \\ \midrule
\multicolumn{4}{c}{\emph{Unseen split}} \\ \midrule
SD v1.4 & 60.1 & 17.3 & 28.7  \\
SD v2.1 & 64.0  & 28.4 & 44.4 \\
SD$_{SR4G}$ v1.4 & 68.9 & 23.7 & 34.4 \\ 
SD$_{SR4G}$ v2.1 & \textbf{69.4} & 29.4 & 42.4 \\ \bottomrule
\end{tabular}
}
\caption{Results obtained for the \emph{main} and \emph{unseen} splits of SR4G. Base models SD v1.4 and v2.1 are shown alongside with fine-tuned SD$_{SR4G}$ models. } 
\label{tab:results}
\end{table}

Table \ref{tab:results} shows the results for our base and fine-tuned models for both SR4G splits, with the best results according to the main comparison metric in bold.  

\textbf{Main split: } We observe that the SD$_{SR4G}$ models improve all metrics respect to the base SD models, increasing both object and spatial relation generation capabilities considerably. These results are in line with our initial hypothesis, proving that the exposure to image-caption pairs with explicit spatial relations improves spatial relation generation. Our results show that SD$_{SR4G}$ v1.4 and v2.1 have almost the same spatial capabilities, but v2.1 excels for object rendering. Notice that the differences of the base SD models are much bigger. 

\textbf{Unseen split:} To analyse whether the improvements of SD$_{SR4G}$ on the \emph{main} split come from learning specific correlations between pairs of objects, or between objects and spatial relations, we check the results on the \emph{unseen} split. The \emph{unseen} split uses different objects in train and test, and it is thus designed to decouple objects from spatial relations, allowing us to focus on the performance for spatial relations in isolation. In Table \ref{tab:results}, we see that both versions of SD$_{SR4G}$ consistently improve the VISOR$_{\mathrm{Cond}}$ and VISOR metrics over the base SD systems, also for the \emph{unseen} split. It is specially interesting that VISOR$_{\mathrm{Cond}}$, which is not influenced by object accuracy, is almost the same as for the \emph{main} split. That means that our models are generalizing to unseen objects during the fine-tuning step. The behaviour of both versions is very similar to the \emph{main} split.

\textbf{Image quality:} As we are using synthetic captions to train, we make sure that the image generation capabilities of these models do not worsen over training. Therefore, we monitor the Fréchet Inception Distance (FID) \cite{heusel2017gans} between the model's generated images from human annotated captions (retrieved from the COCO 2017 validation split) and their respective real images. During all of our experiments FID values have been constant and have not worsen after training. A random set of examples can be seen in Figure \ref{fig:qualAnalysis}.

\begin{table}[t]
\centering
\resizebox{\linewidth}{!}{
\begin{tabular}{lrccc}
\toprule
Model & Par. & {\footnotesize VISOR$_{\mathrm{Cond}}$} $\uparrow$ & {\footnotesize	VISOR} $\uparrow$ & {\footnotesize OA} $\uparrow$ \\ \midrule
\multicolumn{5}{c}{\emph{Main split}} \\ \midrule
LayoutGPT & 8.1B & 64.7 & 24.7 & 38.1 \\
VPGen & 14.1B & 67.7 & 34.5 & 51.0 \\
SD v2.1 & 1.3B & 64.0 & 27.4 & 42.8 \\
SD$_{SR4G}$ v2.1 & 1.3B & \textbf{69.5} & 31.7 & 45.6 \\ \midrule
\multicolumn{5}{c}{\emph{Unseen split}} \\ \midrule
LayoutGPT & 8.1B & 64.7 & 24.7 & 38.1 \\
VPGen $\dagger$ & 14.1B & 68.4 & 37.0 & 54.1 \\ 
SD v2.1 & 1.3B & 64.0 & 28.4 & 44.4 \\
SD$_{SR4G}$ v2.1 & 1.3B & \textbf{69.4} & 29.4 & 42.4 \\
\bottomrule
\end{tabular}
}
\caption{Comparison to the state of the art, including model size for both splits. $\dagger$ VPGen is contaminated, as it was trained on layouts containing spatial triplets that appear in our test split. }
\label{tab:sota}
\end{table}

\subsection{Comparison with the state of the art}
\label{sec:sota}

We also compare against two recent state-of-the-art pipeline models: LayoutGPT and VPGen. The backbone Large Language Model (LLM) of VPGen has already been fine-tuned for layout generation,\footnote{They use three different datasets to obtain caption-layout pairs to fine-tune the LLM: Flickr30K entities \cite{plummer2015flickr30k}, COCO instances 2014 \cite{lin2014microsoft}, and PaintSkills \cite{cho2023dall}.} so we use VPGen with no further adaptation. Note that the layout generation module of VPGen has been trained on COCO, and thus contains the objects underlaying our test sets. In the case of LayoutGPT, adaptation is performed with in-context learning. We thus define a set of instances that will be used as in-context examples to condition the 7B parameter Llama-2 LLM. For this purpose, we randomly extract 400 caption-layout pairs per different relation from our SR4G dataset, and build a set of 5.6k instances of caption-layout pairs. For inference, $k=8$ examples are chosen by computing the CLIP-based similarity \cite{radford2021learning} between the input caption and the set of in-context examples, retrieving the top-$k$ most similar examples and using them to condition the model to generate the proper layout.

Table \ref{tab:sota} shows the obtained results for both SR4G splits. The same trend is observed, i.e. SD$_{SR4G}$ v2.1 clearly outperforms both state-of-the-art pipeline systems in terms of VISOR$_{\mathrm{Cond}}$, which measures the correctness of the spatial relation when both objects are generated. The improvement is especially important considering that both pipeline systems are significantly larger in terms of parameters, with a more complex architecture involving LLMs, and that both are specifically designed to generate scene layouts. 

The table also shows the two auxiliary metrics, with VPGen obtaining the best results for object accuracy and VISOR. That is expected, since VPGen has been trained specifically for object generation, and VISOR is calculated over all the recognised objects. In fact, the better VISOR results are only due to better object accuracy, as our method produces better spatial configurations after factoring out object accuracy from VISOR (VISOR$_{\mathrm{Cond}}$). Also note the contamination issue for the \emph{unseen} split, as the text-to-layout step of VPGen has been fine-tuned on COCO. This implies that VPGen has seen text-layout pairs using the entire set of objects, having been trained on all the objects in our test set.

\section{Analysis}

\begin{table}[t]
    \centering
    \resizebox{0.9\linewidth}{!}{
    \begin{tabular}{clcc} \toprule
        Type & Relation & Main Split & Unseen Split \\ \midrule
        \parbox[t]{2mm}{\multirow{4}{*}{\rotatebox[origin=c]{90}{Projective}}} & \emph{Left of} & 70.3 (+7.0) & 69.8 (+8.8) \\
        & \emph{Right of} & 72.4 (+8.0) & 67.9 (+3.9) \\
        & \emph{Above} & 72.0 (+4.5) & 70.4 (+2.2) \\
        & \emph{Below} & 71.4 (+4.5) & 70.3 (+2.8) \\ \midrule
        \parbox[t]{2mm}{\multirow{4}{*}{\rotatebox[origin=c]{90}{Topological}}} & \emph{Overlapping} & 86.9 (-4.9) & 84.0 (-5.2)  \\
        & \emph{Separated} & 79.5 (+17.0) & 84.8 (+18.5) \\ 
        & \emph{Surrounding} & 29.8 (+2.3) & 21.7 (-2.1) \\
        & \emph{Inside} & 43.4 (-7.4) & 39.2 (-6.4) \\ \midrule
        \parbox[t]{2mm}{\multirow{6}{*}{\rotatebox[origin=c]{90}{Scale}}} & \emph{Taller} & 71.2 (+1.6) & 75.6 (+5.0) \\
        & \emph{Shorter} & 67.5 (+8.5) & 69.0 (+11.9) \\
        & \emph{Wider} & 71.6 (+4.3) & 73.0 (+6.9) \\
        & \emph{Narrower} & 69.3 (+9.3) & 67.1 (+5.0) \\
        & \emph{Larger} & 71.5 (+0.5) & 74.7 (+1.9) \\
        & \emph{Smaller} & 65.2 (+12.7) & 63.3 (+13.5) \\ \bottomrule
    \end{tabular}
    }
    \caption{VISOR$_\mathrm{Cond}$ values per relation obtained by SD$_{SR4G}$ v2.1. The difference in VISOR$_\mathrm{Cond}$ between SD v2.1 and fine-tuned SD$_{SR4G}$ is given in brackets.
    }
    \label{tab:visorRelation}
\end{table}

We show an extensive analysis of the consequences of fine-tuning on SR4G, covering performance per relation, biases for opposite relations, performance by frequency of triplets and qualitative examples.

\subsection{Analysing performance per relation}
\label{sec:analysisRel}

In Table \ref{tab:visorRelation} we show VISOR$_{\mathrm{Cond}}$ values per spatial relation for SD$_{SR4G}$ v2.1 (our best model), both in the \emph{main} and \emph{unseen} splits. 

First, we observe that all projective relations significantly improve for both splits. The improvement is bigger for \emph{left of} and \emph{right of}. That might be due to random horizontal flips applied only to the images during the training of SD models, which are expected to damage the model's ability to correctly learn those relations.

\begin{figure}[t]
    \centering
    \includegraphics[width=\linewidth]{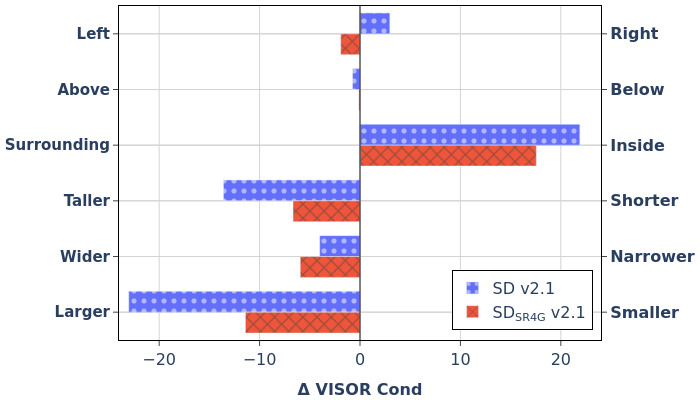}
    \caption{The horizontal axis depicts the difference of VISOR$_\mathrm{Cond}$ values between relation pairs with opposing meanings defined on each side of the vertical axis. Results for SD and SD$_{SR4G}$ v2.1 on the \emph{unseen} split.}
    \label{fig:relationBias}
\end{figure}

Topological relations show a more variable behaviour. In the case of \emph{separated}, our unique topological relation that does not involve generating overlapping objects, SD$_{SR4G}$ is capable of improving its performance by up to 18.5 points VISOR$_\mathrm{Cond}$. However, for \emph{overlapping}, fine-tuning is not helpful. SD v2.1 already knows how to generate images with the \emph{overlapping} relation, achieving VISOR$_\mathrm{Cond}$ values of 91.8 and 89.2 in both test splits. On the other hand, \emph{surrounding} and \emph{inside} seem to be especially hard. The VISOR$_\mathrm{Cond}$ values are low for the SD model and fine-tuning even makes them worse (especially for \emph{inside}). This is a limitation of our current approach, and different training strategies must be explored to tackle this issue.

Finally, SD$_{SR4G}$ improves for all scale relations. It is curious to observe that \emph{taller}, \emph{wider} and \emph{larger} perform better than their opposites, even though the improvements over the base SD model are more modest. That suggests that the base SD model might have a bias towards those spatial relations.

\begin{figure*}[t]
    \begin{minipage}{0.48\textwidth}
        \centering
        \includegraphics[width=\linewidth]{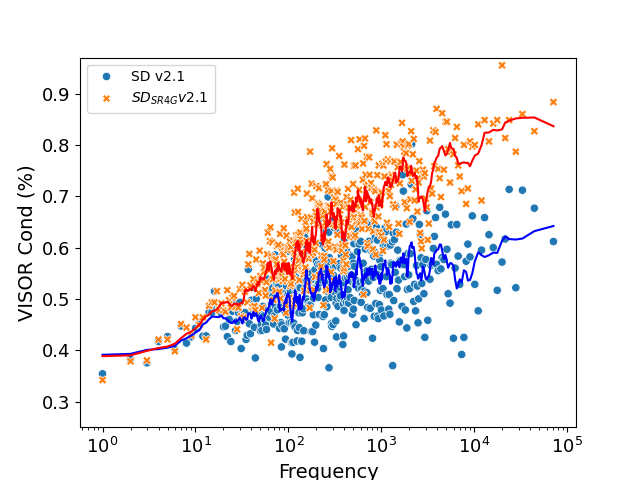}
        \subcaption{Results using \emph{main} splits.}
        \label{fig:correlationOverlap}
    \end{minipage}\hfill
    \begin{minipage}{0.48\textwidth}
        \centering
        \includegraphics[width=\linewidth]{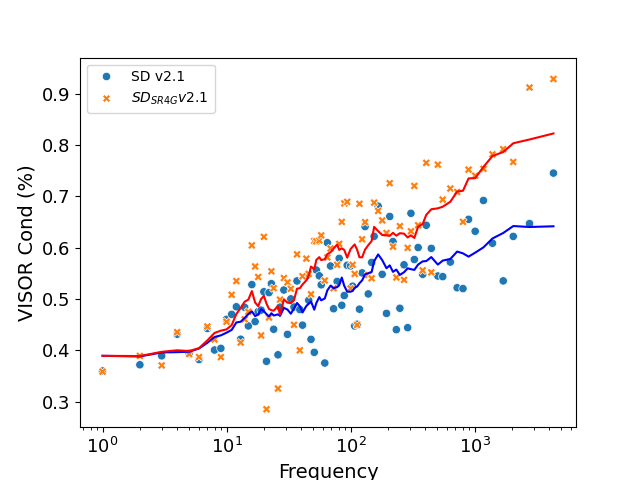}
        \subcaption{Results using \emph{unseen} splits.}
        \label{fig:correlationObjsplit}
    \end{minipage}
    \caption{Correlation between the frequency of SR4G triplets in COCO training instances (shown in the logarithmic horizontal axis) and their respective VISOR$_\mathrm{Cond}$ results for SD v2.1 and SD$_{SR4G}$ v2.1. Triplets are grouped by frequency for visibility.}
    \label{fig:correlations}
\end{figure*}

\subsection{Analysing biases for opposite relations}
\label{sec:analysisPairs}

Most of our relations have an opposite relation, i.e., \emph{right of} is the opposite of \emph{left of}. There are a total of six pairs of opposites in our relation set, which are listed in Figure \ref{fig:relationBias} along with the difference in performance for these pairs before and after fine-tuning using the \emph{unseen} split.

We want to see whether performance biases between opposites are reduced by fine-tuning. Figure \ref{fig:relationBias} shows strong preferences of our base model SD v2.1 (in Appendix \ref{sec:appLAION}, we show that those differences are correlated with the rate of appearance of each relation in the pretraining dataset of the SD models). We can also observe that SD$_{SR4G}$ v2.1 significantly reduces the difference in VISOR$_\mathrm{Cond}$ between all relation pairs (except for \emph{wider} and \emph{narrower}), showing that fine-tuning reduces the inherent biases of the base model.

\subsection{Performance by frequency of triplets}
\label{sec:analysisFreq}

As SR4G is derived from natural images, some triplets are more frequent than others. To measure how the frequency of training triplets affects the results of our fine-tuned models, in Figure \ref{fig:correlations}, we depict the VISOR$_{\mathrm{Cond}}$ values of SD v2.1 and SD$_{SR4G}$ v2.1 depending on the frequency of each triplet in the COCO training set.

Figure \ref{fig:correlationOverlap} shows the results for the \emph{main} split. In this case, the image generator has seen test triplets during training and, as expected, the more frequent these triplets, the greater the improvement after the fine-tuning. We can also observe that, even though SD models have not seen COCO images before, its performance is correlated with our computed frequencies.

On the other hand, Figure \ref{fig:correlationObjsplit} shows a similar plot when training and evaluating on the \emph{unseen} split. We observe similar correlations as in Figure \ref{fig:correlationOverlap} with both models. However, now we are evaluating on images generated from unseen triplets composed by objects that have not been seen during fine-tuning. Therefore, these results show that it is easier to transfer what is learnt to the most common triplets, even though we have not trained on them.

\subsection{Qualitative Analysis}
\label{sec:analysisQual}

\begin{figure*}
    \centering
    \includegraphics[width=\textwidth]{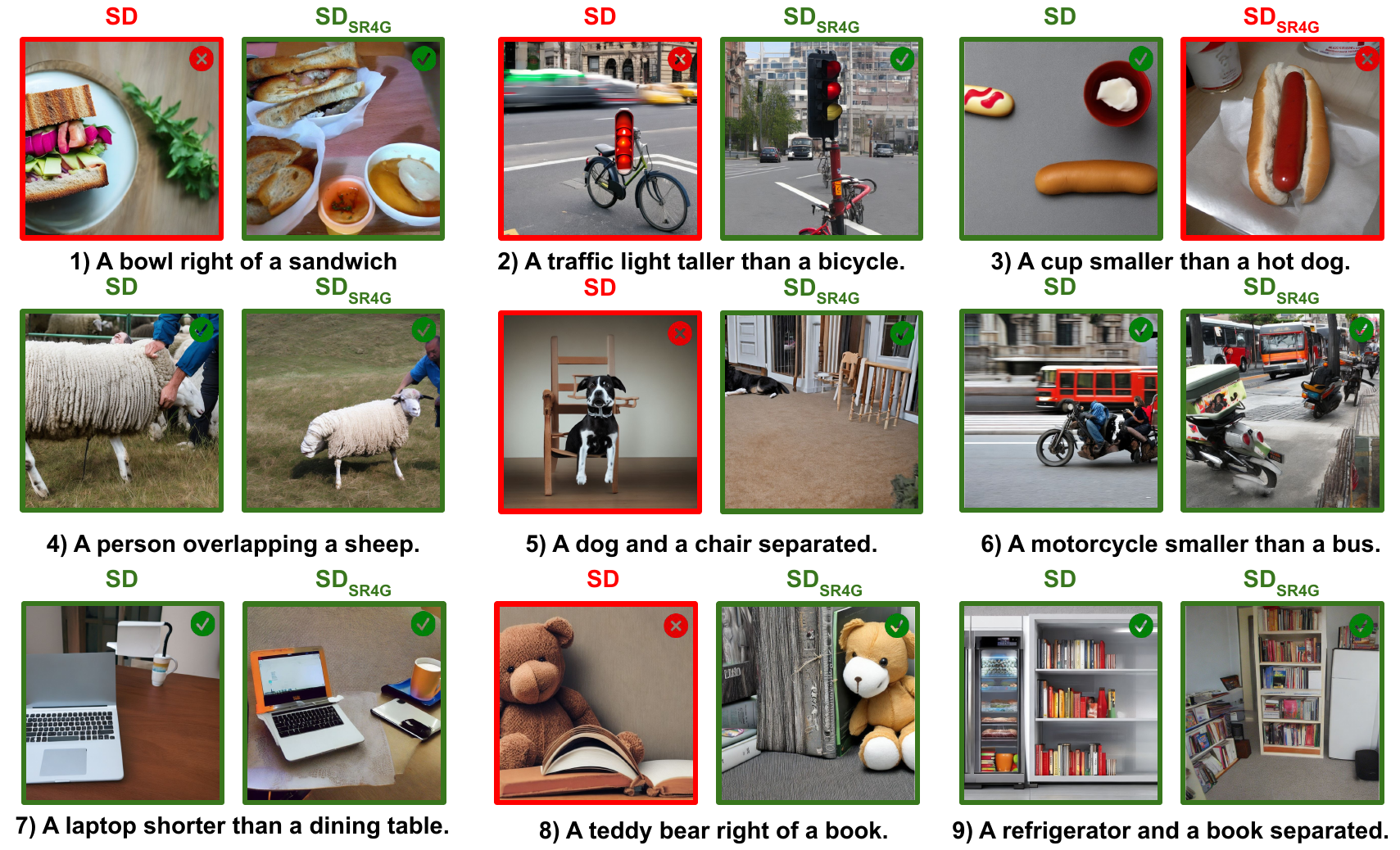}
    \caption{Image generation examples by SD v2.1 and SD$_{SR4G}$ v2.1 fine-tuned on the \emph{main} split. Following our relation-specific heuristics, if the relation in the caption is correctly depicted, we indicate this with a green tick. Otherwise, there is a red cross in the top-right corner of the image.}
    \label{fig:qualAnalysis}
\end{figure*}

In order to visualize and  qualitatively evaluate the generated images, we take SD v2.1 and SD$_{SR4G}$ v2.1 fine-tuned on the \emph{main} split. We discard the most common and uncommon spatial triplets. The rationale is that the most common triplets often contain easy-to-generate relations (e.g.,~$\langle$\textit{truck, larger, dog}$\rangle$) as generating both objects is enough to instantiate the relation itself, whereas the least frequent ones do not seem natural and would not be used in a prompt (e.g.,~$\langle$\textit{bus, shorter, traffic light}$\rangle$). Therefore, we randomly pick triplets that occur between 100 and 1,000 times in COCO annotations (we obtain that range from the frequency analysis in Figure \ref{fig:correlations}). We start generating images using random captions. We keep the first nine image pairs where both objects are generated correctly. Those nine pairs can be found in Figure \ref{fig:qualAnalysis}, where we also indicate whether the spatial relation in the caption is depicted correctly or not.

Some of the captions of Figure \ref{fig:qualAnalysis} describe \emph{easy} spatial relations, such as number 2, 3, 6, 7 and 9, where usually, if the correct objects are generated, the relation is also correct. SD$_{SR4G}$ generates those relations correctly, except for 3, which we denoted as a failure because the cup is not fully visible (the decision is arguable). SD fails for 2, rendering the traffic light very oddly. Captions 1, 4, 5 and 8 are more demanding: SD$_{SR4G}$ correctly depicts all the relations (\emph{right of} twice, \emph{overlapping} and \emph{separated}), but SD fails for 1, 5 and 8. The failures are interesting: for 1 and 8, the spatial relations of the captions might not be the most typical ones in natural images, and SD struggles. However, for 5 it should be very common to see dogs and chairs separated, but SD does not follow the caption, which suggests that the relation \emph{separated} is not known to SD.

\section{Conclusions}

In this work we define a dataset generation pipeline to build synthetic captions containing explicit spatial relations from COCO images and annotations. Fine-tuning diffusion models with these image-caption pairs outperforms the original diffusion models and also surpasses state-of-the-art pipeline models for spatial relation generation. We find that SD$_{SR4G}$ generalizes to unseen objects during fine-tuning. Further analysis shows that SD$_{SR4G}$ learns to better depict projective and scale relations, reduces the bias that the original model has for opposite relations, and generalizes better to spatial triplets that are more frequent in real images.

As future work, we plan to expand our relation set to include depth information with relations such as \emph{in front of} and \emph{behind}. We would also like to explore new ways to collect and annotate natural captions with spatial relations and evaluate state-of-the-art models with them. 

\section{Limitations}
SR4G only contains captions in English, which limits its usage for non-English languages. To make it multi-lingual, caption generation scripts should be modified. On the other hand, SR4G is focused on unambiguous spatial relations defined over bounding box information, since they can be generated and evaluated automatically using off-the-shelf object detectors and heuristic rules. In that sense, orientation relations are discarded, even though their analysis is very interesting. Finally, we focus on 2D spatial relations. To introduce 3D relations should also be possible, using off-the-shelf depth estimation systems for images.   

\bibliography{anthology,custom}

\appendix

\section{Details on SR4G Dataset}
\label{sec:appSR4G}

In this appendix, we give more details about our \emph{main} and \emph{unseen} splits, as well as defining our hand designed templates and heuristics used to determine whether an image contains a given spatial relation between two objects.

\subsection{Hand designed templates}
\label{sec:appSR4G-Templates}

The templates we use to generate captions from spatial triplets are shown in Table \ref{tab:relationTemplates}. As can be seen, those templates are designed to be as simple as possible, omitting attributes and verbs and focusing only on the objects and their spatial relation. This is very important to analyse spatial understanding in isolation.

\begin{table}[t]
    \centering
    \begin{tabular}{ccc}
        \toprule
        Type & Relation & Template \\ \midrule
        \parbox[t]{2mm}{\multirow{4}{*}{\rotatebox[origin=c]{90}{Projective}}} & \emph{Left of} & $\langle$A$\rangle$ to the left of $\langle$B$\rangle$.  \\
        & \emph{Right of} & $\langle$A$\rangle$ to the right of $\langle$B$\rangle$.  \\
        & \emph{Above} & $\langle$A$\rangle$ above $\langle$B$\rangle$. \\
        & \emph{Below} & $\langle$A$\rangle$ below $\langle$B$\rangle$. \\ \midrule
        \parbox[t]{2mm}{\multirow{4}{*}{\rotatebox[origin=c]{90}{Topological}}} & \emph{Overlapping} & $\langle$A$\rangle$ overlapping $\langle$B$\rangle$.  \\
        & \emph{Separated} & $\langle$A$\rangle$ and $\langle$B$\rangle$ separated.  \\ 
        & \emph{Surrounding} &  $\langle$A$\rangle$ surrounding $\langle$B$\rangle$. \\
        & \emph{Inside} & $\langle$A$\rangle$ inside of $\langle$B$\rangle$.\\ \midrule
        \parbox[t]{2mm}{\multirow{6}{*}{\rotatebox[origin=c]{90}{Scale}}} & \emph{Taller} & $\langle$A$\rangle$ taller than $\langle$B$\rangle$. \\
        & \emph{Shorter} & $\langle$A$\rangle$ shorter than $\langle$B$\rangle$. \\
        & \emph{Wider} & $\langle$A$\rangle$ wider than $\langle$B$\rangle$. \\
        & \emph{Narrower} & $\langle$A$\rangle$ narrower than $\langle$B$\rangle$. \\
        & \emph{Larger} &  $\langle$A$\rangle$ larger than $\langle$B$\rangle$. \\
        & \emph{Smaller} &  $\langle$A$\rangle$ smaller than $\langle$B$\rangle$. \\ \bottomrule
    \end{tabular}
    \caption{Templates used to generate synthetic captions.} 
    \label{tab:relationTemplates}
\end{table}

\subsection{Heuristic rules}
\label{sec:appSR4G-Rules}

We use heuristic rules to both build the dataset and evaluate the generated images. Assuming the spatial triplet $\langle l_{s}, r, l_{o}\rangle$ and the bounding boxes of its objects $bb_{s}$ and $bb_{o}$ that appear in an image, we define the heuristic rule $f_r$ of relation $r$ to determine whether the triplet is fulfilled in the image or not. We set $bb_{i} = \{x^0_i , y^0_i , x^1_i , y^1_i\}$ by defining the top-left $\{x^0_i, y^0_i\}$ and bottom-right coordinates $\{x^1_i, y^1_i\}$ of the bounding-box (\emph{bbox}).

For \emph{left of}, \emph{right of}, \emph{above} and \emph{below}, we follow the heuristic rules defined in \cite{gokhale2023benchmarking}, by computing the centroid of each \emph{bbox} $c_i = \{x^c_i, y^c_i\}$ and comparing their corresponding coordinates. 

As we expand to 10 more relations, we follow the rules described in \cite{johnson2018image}. In our scale relations we compare either the height (\emph{taller} and \emph{shorter}), width (\emph{wider} and \emph{narrower}) or area (\emph{larger}, \emph{smaller}) difference between both \emph{bboxes}. In the cases of \emph{surrounding} and \emph{inside}, we check whether $bb_o$ is contained in $bb_s$ or vice versa. Finally, using the Intersection over Union (IoU) of both \emph{bboxes}, we say that both objects are \emph{separated} if their IoU is 0, and \emph{overlapping} if their IoU is positive.

\subsection{Main and Unseen Splits}
\label{sec:appSR4G-Splits}

\begin{table}[t]
    \centering
    \begin{tabular}{p{\linewidth}}
        \toprule
        \multicolumn{1}{c}{$O_{\mathrm{Train}}$} \\ \midrule
        \emph{person}, \emph{car}, \emph{motorcycle}, \emph{airplane}, \emph{train}, \emph{boat}, \emph{fire hydrant}, \emph{bench}, \emph{bird}, \emph{elephant}, \emph{bear}, \emph{giraffe}, \emph{handbag}, \emph{tie}, \emph{snowboard}, \emph{baseball bat}, \emph{baseball glove}, \emph{surfboard}, \emph{cup}, \emph{knife}, \emph{spoon}, \emph{apple}, \emph{sandwich}, \emph{orange}, \emph{broccoli}, \emph{carrot}, \emph{pizza}, \emph{donut}, \emph{chair}, \emph{couch}, \emph{potted plant}, \emph{bed}, \emph{dining table}, \emph{toilet}, \emph{laptop}, \emph{mouse}, \emph{remote}, \emph{keyboard}, \emph{oven}, \emph{sink}, \emph{book}, \emph{clock}, \emph{teddy bear}, \emph{hair drier}, \emph{toothbrush} \\ \midrule
        \multicolumn{1}{c}{$O_{\mathrm{Val}}$} \\ \midrule
        \emph{umbrella}, \emph{cake}, \emph{tv}, \emph{refrigerator}, \emph{vase} \\ \midrule
        \multicolumn{1}{c}{$O_{\mathrm{Test}}$} \\ \midrule 
        \emph{bicycle}, \emph{bus}, \emph{truck}, \emph{traffic light}, \emph{stop sign}, \emph{parking meter}, \emph{cat}, \emph{dog}, \emph{horse}, \emph{sheep}, \emph{cow}, \emph{zebra}, \emph{backpack}, \emph{suitcase}, \emph{frisbee}, \emph{skis}, \emph{sports ball}, \emph{kite}, \emph{skateboard}, \emph{tennis racket}, \emph{bottle}, \emph{wine glass}, \emph{fork}, \emph{bowl}, \emph{banana}, \emph{hot dog}, \emph{cell phone}, \emph{microwave}, \emph{toaster}, \emph{scissors} \\ \bottomrule
    \end{tabular}
    \caption{Objects used in train, val and test sets of our \emph{Unseen split}.} 
    \label{tab:unseenSplitObjects}
\end{table}

Table \ref{tab:unseenSplitObjects} shows the sets of objects used for training, validation and test in the \emph{unseen} split, which we refer to as $O_{\mathrm{train}}$, $O_{\mathrm{val}}$ and $O_{\mathrm{test}}$, respectively.

There are few combinations that can be built with $O_{\mathrm{val}}$ for validation in the \emph{unseen} split, so we select triplets that contain one object from $O_{val}$ at least once and do not contain any object that is set aside for the test split. In other words, there are up to $(2 \cdot |O_{\mathrm{train}}| \cdot |O_{\mathrm{val}}| + {|O_{\mathrm{val}}| \choose 2}) \cdot 14 = 6,580$ triplets that fulfil this rule (around 5,326 that naturally occur in the COCO dataset). 

Validation is computationally costly in both splits, as several images have to be generated to compute the evaluation metrics defined in Section \ref{sec:eval}. Preliminary experiments showed that generating just 10k images is enough to get consistent results. Thus, we randomly selected 2.5k spatial captions for the validation splits for both \emph{main} and \emph{unseen} splits (as we generate 4 images per caption).

\section{Training settings}
\label{sec:appTrain}

\begin{table}[t]
    \centering
    \begin{tabular}{lc}
        \toprule
        Hyperparameter & Value \\ \midrule
        Training steps & 100k \\
        Batch size & 64   \\
        Learning Rate & $10^{-5}$   \\
        Optimizer & AdamW  \\
        Adam $\beta_1$ & 0.9 \\
        Adam $\beta_2$ & 0.999 \\
        Adam $\epsilon$ & $10^{-8}$ \\
        Weight decay & 0.01   \\
        Mixed-precision & bf16   \\ \bottomrule
    \end{tabular}
    \caption{Fine-tuning hyperparameters of the diffusion models.}
    \label{tab:hyperparameters}
\end{table}

\textbf{Hyperparameters}: In Table \ref{tab:hyperparameters} we define the hyperparameters used for training. Learning rate and optimizer parameters are the ones used during the pretraining of SD models, the other listed hyperparameters have been adapted to our available infrastructure. We also take advantage of Exponential Moving Average \cite{DBLP:journals/corr/KingmaB14} to update the parameters of the models with an AdamW optimizer \cite{loshchilov2018decoupled} and we do not use any learning-rate scheduler. We do validation runs every 5k steps and do not set any early-stopping mechanism.

\textbf{GPU usage:} Due to different memory needs, we use 2 and 4 NVIDIA A100 GPUs to fine-tune SD v1.4 and SD v2.1 models, respectively. In both cases we use an effective batch size of 64 by changing the amount of instances assigned to each GPU. Each of our fine-tunings need 3 days to be completed.

\textbf{Data augmentation}: During training we apply random horizontal flips and random crops to our images as a data augmentation strategy (resulting in $I^{*}$ and $O^{j}$). Note that, random horizontal flips are common during the training of text-to-image models. This implies that spatial relations, such as \emph{left of} and \emph{right of}, can not be learnt correctly (as captions are not transformed according to those flips). Nevertheless, in our case we apply the same transformations to \emph{bboxes}, which are used to generate captions synthetically, keeping this data augmentation strategy while maintaining the generated caption's spatial correctness.

Random crops might reduce the number of objects in $O_{I^{*}}$. If there are less than two objects after a given crop, we redo it up to $max\_iter$ times until there are at least two objects in the image. 

We also define the hyperparameter $k$ as the number of captions that can be concatenated to build the image-caption pairs built during training. Table \ref{tab:tripletAblation} shows the results obtained by concatenating $k \in  \{1, \ldots, 5\}$ captions. We observe that $k = 2$ obtains the best results, and we use this value of $k$ during our entire work.

\begin{table}[t]
    \centering
    \resizebox{\linewidth}{!}{
        \begin{tabular}{cccc}
            \toprule
            Nº Captions & VISOR$_{\mathrm{Cond}}$ $\uparrow$ & VISOR $\uparrow$ & OA $\uparrow$  \\ \midrule
            1 & 68.1 & 26.5 & 38.9 \\
            2 & \textbf{69.4} & 27.4 & 39.5 \\
            3 & 67.7 & 27.1 & 40.0 \\
            4 & 63.7 & 21.9 & 34.3 \\
            5 & 63.0 & 22.9 & 36.3 \\ \bottomrule
        \end{tabular}
    }
    \caption{We fine-tune SD v1.4 in the \emph{main} split concatenating different amounts of captions in the input. These results correspond to the validation set of our \emph{main} split.}
    \label{tab:tripletAblation}
\end{table}

\section{Evaluation settings}
\label{sec:appEval}

The evaluation metrics used in this paper use an object detector to determine whether objects are generated correctly and where are located in the image. Following \cite{gokhale2023benchmarking}, we use OWL-ViT, an open-vocabulary object detector that uses a CLIP \cite{radford2021learning} backbone with a ViT-B/32 transformer architecture \cite{zhai2022scaling}. We also set 0.1 as the confidence threshold of OWL-ViT, which determines how sure the model must be for a given region of the image to contain a specific object.

As an open-vocabulary object detector, OWL-ViT takes as input the objects we want to detect and, in order to do so, we use their recommended template ("a photo of a $\langle$OBJ$\rangle$.") instead of the object label alone.

Due to the variability of images generated by Stable Diffusion, we generate 4 images per evaluation caption. Therefore, we generate 10k images per validation and a total of 243.3k and 32.1k images to test each model in the \emph{main} and \emph{unseen} splits, respectively.

\section{LAION Dataset and Spatial Relations}
\label{sec:appLAION}
\begin{figure}[t]
    \centering
    \includegraphics[width=\linewidth]{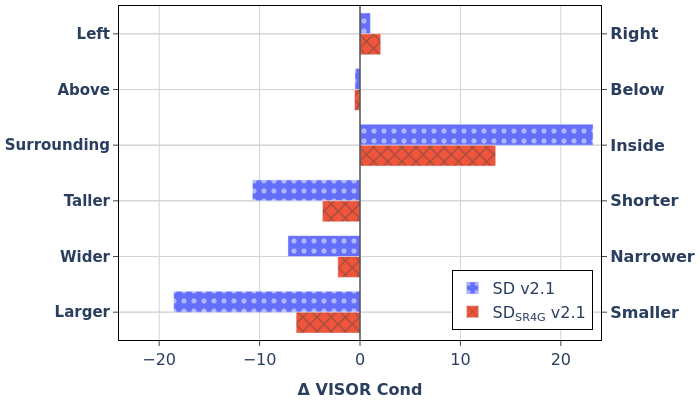}
    \caption{The horizontal axis depicts the difference of VISOR$_\mathrm{Cond}$ values between relation pairs with opposing meanings defined on each side of the vertical axis. These results correspond to SD and SpaD v2.1 trained and evaluated using \emph{main} splits.}
    \label{fig:relationBiasBase}
\end{figure}

Figure \ref{fig:relationBias} shows that Stable Diffusion models have a strong bias towards some spatial relations, preferring \emph{taller} to \emph{shorter}, for instance. To complete those results, we also show the same graphic but in the \emph{main} split, which exhibits a very similar behaviour (Figure \ref{fig:relationBiasBase}). To understand the origin of those biases, we checked the frequency of each spatial relation in the LAION-2B dataset (English subset), used to train SD models. Table \ref{tab:laionRel} shows the appearances of 12 relations, divided in 6 relation pairs with opposite meanings. Every relation has its number of appearances in LAION into brackets. For each opposite relation pair, the first column contains the relation that best works with SD. The third column shows the ratio of appearance between the preferred relation and its opposite (>1 indicates that the preferred relation appears more times in LAION than its opposite relation). The results indicate that there is a clear correlation between the ratio of appearance of a relation and the bias of SD models. The only exception is the \emph{right} and \emph{left} pair, but both appear similar times and the bias towards \emph{right} is very small.

\begin{table}[t]
    \centering
    \resizebox{\linewidth}{!}{
    \begin{tabular}{ccc}
        \toprule
         \multirow{2}{*}{Preferred Rel.} & \multirow{2}{*}{Opposite Rel.} & \multirow{2}{*}{Ratio of Appearance}  \\ 
         & & \\ \midrule
         Right (5M) & Left (5.6M) & 0.91 \\
         Above (1.6M) & Below (0.7M) & 2.47 \\
         Inside (2M) & Surrounding (0.3M) & 7.61 \\
         Taller (49.3K) & Shorter (29.4K) & 1.86 \\
         Wider (54.6K) & Narrower (5.7K) & 9.62 \\
         Larger (0.8M) & Smaller (0.2M) & 3.17 \\ \bottomrule
    \end{tabular}
    }
    \caption{Ratio in which the first relation appears more than the other. The relation in the first column is the preferred one by SD.}
    \label{tab:laionRel}
\end{table}

\end{document}